\documentclass[conference]{IEEEtran}
\usepackage{times}
\usepackage{graphicx} 
\usepackage{algorithm}
\usepackage[utf8]{inputenc}
\usepackage{algpseudocode}
\usepackage{amsmath}

\usepackage[T1]{fontenc}
\let\labelindent\relax
\usepackage[inline]{enumitem}
\usepackage[table,xcdraw]{xcolor}
\usepackage{multirow}
\usepackage{color}
\usepackage{tabularray}
\usepackage{pgfplots}
\usepackage{tikz}
\usepackage{subcaption}
\pgfplotsset{compat=newest}
\usepackage{xcolor}
\usepackage{setspace}
\usepackage{amssymb}
\usepackage{mathtools}
\usepackage{adjustbox}
\usepackage[hidelinks,bookmarks=true]{hyperref}
\usepackage{booktabs}
\usepackage[numbers]{natbib}
\usepackage{multicol}

\begin{document}

\title{Kinematic Model Optimization via Differentiable Contact Manifold for In-Space Manipulation} 

\author{Abhay Negi, Omey M. Manyar, Satyandra K. Gupta \\
    \small Realization of Robotic Systems Lab, \\ \small University of Southern California, Los Angeles, CA, USA \\ 
    \small Email: abhay.negi@usc.edu
}

\maketitle

\IEEEpeerreviewmaketitle

\section{Introduction}
 
\noindent Robotic manipulation in space is emerging as a key capability for enabling critical operations such as orbital debris removal and in-space servicing, assembly, and manufacturing (ISAM) \cite{zhang_modularity_2023, nanjangud_towards_2024, manz_robotic_nodate, arney_-space_2024, mcguire_everybody_2019, daca_concept_2024, wang_robust_2022, karumanchi_payloadcentric_2018, mccormick_remora_2018}. A critical requirement for these systems is the ability to perform \textit{precise contact-rich manipulation} tasks \cite{papadopoulos_robotic_2021, roa_autonomous_2024}. However, achieving precision in the space environment is particularly challenging due to thermal effects on mechanical and sensing components \cite{wang_space_2021}. Specifically, thermal deformation of a manipulator's links and temperature-dependent bias and noise in joint encoder measurements introduce significant uncertainty into the kinematic model. Due to the extreme thermal variations—often ranging over 100$^{\circ}$C—in the space environment, and the compounding nature of uncertainty in serial manipulators, end-effector pose errors may accumulate to several millimeters and degrees.

To mitigate the impact of kinematic parameter uncertainty, one strategy is to use closed-loop feedback within the task's pose space. Visual servoing \cite{amaya-mejia_visual_2024, zhao_minimum_2020, abdul_hafez_reactionless_2017, alepuz_direct_2016, yang_ground_2014, inaba_visual_2003} and force-based control \cite{su_robot_2022, sy_horng_ting_review_2025, nottensteiner_towards_2021, papadopoulos_robotic_2021, mitsioni_safe_2021} are among the most widely used techniques for this purpose. Nevertheless, the presence of kinematic parameter uncertainty limits the achievable accuracy in end-effector pose; additionally, dynamic lighting and thermal variations in the space environment may further introduce errors in computer vision algorithms and force-torque sensors, and exacerbate pose errors. Thus, parameter estimation is necessary to achieve reliable and safe in-space manipulation. 

Kinematic parameter calibration techniques have been developed, but typically require external sensing such as laser sensors \cite{cho_screw_2019, jiang_stable_2021} or displacement sensors \cite{noauthor_robust_nodate, li_searching_nodate}. Alternatively, techniques such as single point repeatability calibration using a probe tip \cite{liu_structural_2022} enables calibration without external sensing. While these techniques offer high accuracy, they incur operational downtime and may introduce risks associated with interrupting contact-rich tasks to perform calibration. This limitation becomes more severe in dynamic scenarios where frequent re-calibration is necessary and switching between calibration and task execution can be unsafe or impractical. 

This paper presents a novel approach for estimating kinematic parameters using only proprioceptive sensing and force-based contact detection while executing a contact-rich manipulation task. Specifically, we estimate manipulator link lengths and joint encoder biases from encoder measurements during peg-in-hole assembly. The method leverages the contact manifold - the set of $SE(3)$ poses between two rigid bodies at which contact occurs. The structure of the contact manifold serves as a source of information which is used to perform kinematic parameter estimation. The contributions of this paper are twofold: (1) a differentiable, learning-based model of the contact manifold, and (2) an optimization-based algorithm that estimates manipulator link strain and joint encoder biases using only joint encoder measurements. The interpretability of this method, which does not rely on vision or precise force sensing, makes it well-suited for robust and safe manipulation in the challenging conditions of space.

\section{Problem Formulation} 

\noindent We consider a serial robotic manipulator with $n$ revolute joints and $n+1$ links performing a peg-in-hole assembly task. The peg $\mathcal{P}$ is rigidly attached to the manipulator's end effector, while the hole $\mathcal{H}$ is fixed with respect to the base. We assume $\mathcal{P}$ and $\mathcal{H}$ have known geometries and exhibit minimal thermal deformation such that the change in the contact manifold $\mathcal{M}$ is negligible. Although the current formulation does not explicitly model thermal effects on the assembly geometries, future work will address this by incorporating temperature-dependent contact manifold adaptation. 

To characterize thermal deformation of the manipulator, let $l_0$ denote the nominal link lengths, and let $l^*$ denote the current, temperature-affected link lengths. Assuming uniform temperature and a constant thermal expansion coefficient for each link, we model the resulting strain as a scalar parameter $r$, such that the link lengths satisfy the relation $l^* = (1+r) l_0$. 

The manipulator's true angular joint positions are denoted as $q^* \in \mathbb{R}^n$, and the corresponding biased encoder measurements are $q = q^* + b$, where $b$ is a constant bias vector. We define the observation set $\{q\} \in \mathbb{R}^{m \times n}$ as $m$ joint position measurements at which contact is detected between $\mathcal{P}$ and $\mathcal{H}$. Contact is determined to occur when the measured external force magnitude exceeds a threshold, i.e., $|F_{ext}| > \epsilon_f$.

Our goal is to estimate the kinematic parameters $\theta = \{r,b\}$ from the contact observations $\{q\}$. To this end, we utilize the contact manifold, $\mathcal{M}$, which represents the set of peg with respect to hole contact poses $^hT_p$. We define a metric projection function $\mathcal{F}$, which maps an estimated peg with respect to hole contact pose to its nearest point on $\mathcal{M}$. The parameter estimation function is then defined as: $\Psi(\{q\}, \mathcal{M}) \mapsto \theta$. 

\begin{figure*}[ht]
    \centering
    \includegraphics[width=1.0\linewidth]{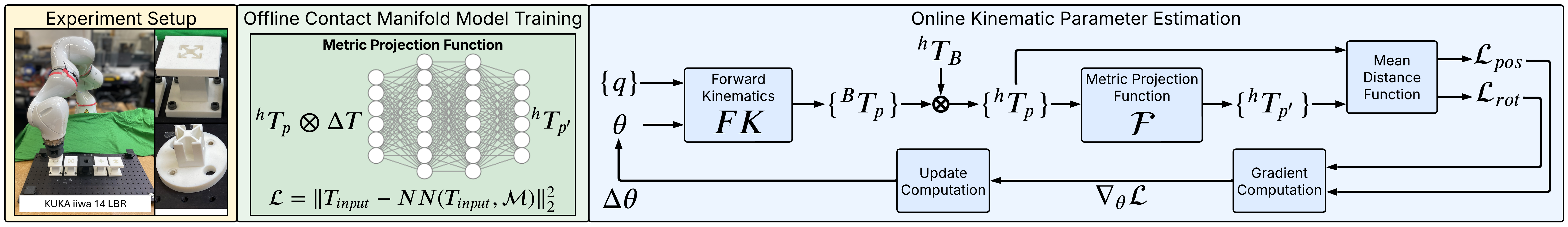}
    \caption{Overview of our methodology. In the offline phase, real-world contact poses are used to train a model, $\mathcal{F}$, that projects nearby points onto the contact manifold. In the online phase, biased encoder measurements at contact are used to estimate the observed peg-hole contact poses. The kinematic parameters are optimally solved for by using gradient-based optimization and minimizing the mean distance between the computed contact poses and their respective projections onto the contact manifold.}
    \label{fig:methodology}   
\end{figure*}

\section{Methodology}

\noindent To solve for kinematic parameters $b$ and $r$,  we propose an approach illustrated in Fig. \ref{fig:methodology}. Our methodology consists of two stages: an offline contact manifold metric projection learning phase and an online kinematic parameter estimation stage. In the offline phase, our goal is to learn a metric projection function, $\mathcal{F}$, which maps poses to the nearest pose on the contact manifold, $\mathcal{M}$. During the online phase, the objective is to align pose estimates from joint position observations with the contact manifold by using gradient-based optimization to solve for the link strain and encoder biases. 

\noindent \textbf{Metric Projection Function}

A metric projection maps each element of a metric space to its nearest element in a designated sub-space. In this work, we train a MLP to approximate a metric projection function mapping poses to the nearest pose on the contact manifold. Each pose is represented as $T = \left[ x,y,z,\alpha,\beta,\gamma\right]^T \in \mathbb{R}^6$. The contact manifold, $\mathcal{M} = \{^hT_p\}$, is constructed from real-world data capturing poses of contact between the peg and hole. Training samples are generated by perturbing poses from $\mathcal{M}$ with uniformly random offsets bounded within $\pm10$ mm and $\pm10^\circ$, i.e. $T_{input} = ^hT^{\mathcal{M}}_p \otimes \Delta T$. The MLP is trained to output, $T_{output}$, minimizing the loss function, $\mathcal{L} = \| T_{input} - NN(T_{input}, \mathcal{M}) \|_2^2$, where $NN(T_{input}, \mathcal{M})$ is the nearest neighbor of $T_{input}$ from the set of poses in $\mathcal{M}$. Assuming small angular perturbations, we treat the manifold as locally Euclidean and use Euclidean distance as a valid $SE(3)$ distance metric. The MLP architecture comprises four hidden layers, each with a dimension of 4096 and ReLU activations. 

\noindent \textbf{Kinematic Parameter Estimation} 

To estimate the link strain and encoder biases, we begin by computing the end-effector pose from the measured joint positions and parameter estimates (initialized to zero) by using forward kinematics. Given the known pose of the hole in the base frame, we compute the peg's pose relative to the hole. Each estimated pose is then projected onto the contact manifold using the metric projection function, $\mathcal{F}$. We compute the mean positional and rotational L1 distances between the estimated poses and their projections. The sum of these distances define the loss function which is minimized using gradient descent to solve for the kinematic parameters. As the forward kinematics, pose transformations, MLP, and distance functions are all differentiable, PyTorch's autograd engine is utilized for efficient gradient computation. 

\section{Results} \label{sec:results}

\noindent We validate our kinematic parameter estimation methodology using contact observations collected from a 7-DOF KUKA LBR iiwa 14 robotic manipulator performing a peg-in-hole assembly task of an extrusion profile geometry with 1.0 mm clearance. We introduce synthetic link strain perturbations of $\pm0.05$ and encoder bias errors of $\pm5^\circ$ into the kinematic model. Over 20 experimental trials, our method achieves a 2.6-fold reduction in link strain error and a 3.37-fold reduction in joint encoder bias error. This error reduction enables sub-mm and sub-degree level accuracy in end-effector pose control. 

\begin{table}[h]
\centering
\caption{Kinematic Parameter Estimation Results}
\label{tab:results}
\begin{tabular}{@{}ll@{}}
\toprule
\textbf{Parameter}     & \textbf{Value}                      \\ \midrule
Link Strain            & $\pm 0.05$                          \\
Encoder Bias           & $\pm5^\circ \cdot \boldsymbol{1}_7$                                  \\
Num. Observations & 3000 (collected at 300 Hz)            \\
Strain MAE             & 0.019 \\
Encoder Bias MAE       & 1.48$^\circ$                        \\ \bottomrule
\end{tabular}
\end{table}

\section{Conclusion} \label{sec:conclusion}

\noindent We presented a contact manifold-based approach for kinematic parameter estimation using only encoder measurements and binary contact detection. The approach is data-efficient, interpretable, and robust, making it well-suited for online in-space manipulator calibration. Our method employs an optimization-based algorithm to estimate link thermal deformation strain and joint encoder biases by regressing contact observations to the contact manifold. Importantly, the metric projection model representing the contact manifold is differentiable - this key feature facilitates continuous and tractable optimization and enables parameter estimation. The differentiability of the model also enables the estimation of other parameters within the forward model, such as the hole pose, and facilitates its use as an external source of information, for instance in camera extrinsic matrix calibration or force-torque sensor calibration. 

In future work, we plan to extend this framework by including additional kinematic parameters and relaxing current modeling assumptions. We also plan to include vision and wrench data to further enhance accuracy and robustness. Lastly, we plan to incorporate uncertainty quantification and provide theoretical guarantees on convergence errors. 

\newpage 

\section*{Acknowledgments}

This work was supported by NASA Space Technology Graduate Research Opportunities Award (80NSSC24K1380).

\bibliographystyle{plainnat}
\bibliography{references}

\end{document}